# Discriminative Probabilistic Models for Relational Data

Ben Taskar
Computer Science Dept.
Stanford University
Stanford, CA 94305
btaskar@cs.stanford.edu

Pieter Abbeel
Computer Science Dept.
Stanford University
Stanford, CA 94305
abbeel@cs.stanford.edu

Daphne Koller
Computer Science Dept.
Stanford University
Stanford, CA 94305
koller@cs.stanford.edu

## Abstract

In many supervised learning tasks, the entities to be labeled are related to each other in complex ways and their labels are not independent. For example, in hypertext classification, the labels of linked pages are highly correlated. A standard approach is to classify each entity independently, ignoring the correlations between them. Recently, Probabilistic Relational Models, a relational version of Bayesian networks, were used to define a joint probabilistic model for a collection of related entities. In this paper, we present an alternative framework that builds on (conditional) Markov networks and addresses two limitations of the previous approach. First, undirected models do not impose the acyclicity constraint that hinders representation of many important relational dependencies in directed models. Second, undirected models are well suited for discriminative training, where we optimize the conditional likelihood of the labels given the features, which generally improves classification accuracy. We show how to train these models effectively, and how to use approximate probabilistic inference over the learned model for collective classification of multiple related entities. We provide experimental results on a webpage classification task, showing that accuracy can be significantly improved by modeling relational dependencies.

## 1 Introduction

The vast majority of work in statistical classification methods has focused on "flat" data – data consisting of identically-structured entities, typically assumed to be independent and identically distributed (IID). However, many real-world data sets are innately relational: hyperlinked webpages, cross-citations in patents and scientific papers, social networks, medical records, and more. Such data consist of entities of different types, where each entity type is characterized by a different set of attributes. Entities are related to each other via different types of links, and the link structure is an important source of information.

Consider a collection of hypertext documents that we want to classify using some set of labels. Most naively, we can use a bag of words model, classifying each webpage solely using the words that appear on the page. However, hypertext has a very rich structure that this approach loses entirely. One document has hyperlinks to others, typically indicating that their topics are related. Each document also has internal structure, such as a partition into sections; hyperlinks that emanate from the same section of the document are even more likely to point to similar documents. When classifying a collection of documents, these are important cues, that can potentially help us achieve better classification accuracy. Therefore, rather than classifying each document separately, we want to provide a form of *collective classification*, where we simultaneously decide on the class labels of all of the entities together, and thereby can explicitly take advantage of the correlations between the labels of related entities.

We propose the use of a joint probabilistic model for an entire collection of related entities. Following the approach of Lafferty (2001), we base our approach on discriminatively trained undirected graphical models, or *Markov networks* (Pearl 1988). We introduce the framework of *relational Markov network (RMNs)*, which compactly defines a Markov network over a relational data set. The graphical structure of an RMN is based on the relational structure of the domain, and can easily model complex patterns over related entities. For example, we can represent a pattern where two linked documents are likely to have the same topic. We can also capture patterns that involve groups of links: for example, consecutive links in a document tend to refer to documents with the same label. As we show, the use of an undirected graphical model avoids the difficulties of defining a coherent generative model for graph structures in directed models. It thereby allows us tremendous flexibility in representing complex patterns.

Undirected models lend themselves well to discriminative training, where we optimize the conditional likelihood of the labels given the features. Discriminative training, given sufficient data, generally provides significant improvements in classification accuracy over generative training (Vapnik 1995). We provide an effective parameter esti-mation algorithm for RMNs which uses conjugate gradient combined with approximate probabilistic inference (belief propagation (Pearl 1988)) for estimating the gradient. We also show how to use approximate probabilistic inference over the learned model for collective classification of multiple related entities. We provide experimental results on a webpage classification task, showing significant gains in accuracy arising both from the modeling of relational dependencies and the use of discriminative training.

## 2 Relational Classification

Consider hypertext as a simple example of a relational domain. A relational domain is defined by a schema, which describes entities, their attributes and relations between them. In our domain, there are two entity types: Doc and Link. If a webpage is represented as a bag of words, Doc would have a set of boolean attributes Doc.$HasWord_k$ indicating whether the word $k$ occurs on the page. It would also have the label attribute Doc.Label, indicating the topic of the page, which takes on a set of categorical values. The Link entity type has two attributes: Link.From and Link.To, both of which refer to Doc entities.

In general, a *schema* specifies of a set of entity types $\mathcal{E} = \{E_1, \ldots, E_n\}$. Each type $E$ is associated with three sets of attributes: content attributes $E.\mathbf{X}$ (e.g. Doc.$HasWord_k$), label attributes $E.\mathbf{Y}$ (e.g. Doc.Label), and reference attributes $E.\mathbf{R}$ (e.g. Link.To). For simplicity, we restrict label and content attributes to take on categorical values. Reference attributes include a special unique key attribute $E.K$ that identifies each entity. Other reference attributes $E.R$ refer to entities of a single type $E' = Range(E.R)$ and take values in $Domain(E'.K)$.

An *instantiation* $\mathcal{I}$ of a schema $\mathcal{E}$ specifies the set of entities $\mathcal{I}(E)$ of each entity type $E \in \mathcal{E}$ and the values of all attributes for all of the entities. For example, an instantiation of the hypertext schema is a collection of webpages, specifying their labels, words they contain and links between them. We will use $\mathcal{I}.\mathbf{X}$, $\mathcal{I}.\mathbf{Y}$ and $\mathcal{I}.\mathbf{R}$ to denote the content, label and reference attributes in the instantiation $\mathcal{I}$; $\mathcal{I}.\mathbf{x}$, $\mathcal{I}.\mathbf{y}$ and $\mathcal{I}.\mathbf{r}$ to denote the values of those attributes. The component $\mathcal{I}.\mathbf{r}$, which we call an *instantiation skeleton* or *instantiation graph*, specifies the set of entities (nodes) and their reference attributes (edges). A hypertext instantiation graph specifies a set of webpages and links between them, but not their words or labels.

The structure of the instantiation graph has been used extensively to infer their importance in scientific publications (Egghe and Rousseau 1990) and hypertext (Kleinberg 1999). Several recent papers have proposed algorithms that use the link graph to aid classification. Chakrabarti *et al.* (1998) use system-predicted labels of linked documents to iteratively re-label each document in the test set, achieving a significant improvement compared to a baseline of using the text in each document alone. A similar approach was used by Neville and Jensen (2000) in a different domain. Slattery and Mitchell (2000) tried to identify directory (or hub) pages that commonly list pages of the same topic, and used these pages to improve classification of university webpages. However, none of these approaches provide a coherent model for the correlations between linked webpages. Thus, they apply combinations of classifiers in a procedural way, with no formal justification.

Taskar *et al.* (2001) suggest the use of *probabilistic relational models (PRMs)* for the collective classification task. PRMs (Koller and Pfeffer 1998; Friedman *et al.* 1999) are a relational extension to Bayesian networks (Pearl 1988). A PRM specifies a probability distribution over instantiations consistent with a given instantiation graph by specifying a Bayesian-network-like template-level probabilistic model for each entity type. Given a particular instantiation graph, the PRM induces a large Bayesian network over that instantiation that specifies a joint probability distribution over all attributes of all of the entities. This network reflects the interactions between related instances by allowing us to represent correlations between their attributes.

In our hypertext example, a PRM might use a naive Bayes model for words, with a directed edge between Doc.Label and each attribute Doc.$HadWord_k$; each of these attributes would have a *conditional probability distribution* $P(\text{Doc}.HasWord_k \mid \text{Doc}.Label)$ associated with it, indicating the probability that word $k$ appears in the document given each of the possible topic labels. More importantly, a PRM can represent the inter-dependencies between topics of linked documents by introducing an edge from Doc.Label to Doc.Label of two documents if there is a link between them. Given a particular instantiation graph containing some set of documents and links, the PRM specifies a Bayesian network over all of the documents in the collection. We would have a probabilistic dependency from each document's label to the words on the document, and a dependency from each document's label to the labels of all of the documents to which it points. Taskar *et al.* show that this approach works well for classifying scientific documents, using both the words in the title and abstract and the citation-link structure.

However the application of this idea to other domains, such as webpages, is problematic since there are many cycles in the link graph, leading to cycles in the induced "Bayesian network", which is therefore not a coherent probabilistic model. Getoor *et al.* (2001) suggest an approach where we do not include direct dependencies between the labels of linked webpages, but rather treat links themselves as random variables. Each two pages have a "potential link", which may or may not exist in the data. The model defines the probability of the link existence as a function of the labels of the two endpoints. In this link existence model, labels have no incoming edges from other labels, and the cyclicity problem disappears. This model, however, has other fundamental limitations. In particular, the resulting Bayesian network has a random variable for each potential link — $N^2$ variables for collections containing $N$ pages. This quadratic blowup occurs even when the actual link graph is very sparse. When $N$ is large (e.g., the set of all webpages), a quadratic growth is intractable. Even more problematic are the inherent limitations on the expressive power imposed by the constraint that the directed graph must represent a coherent generative model over graph structures. The link existence model assumes that the presence of different edges is a conditionally independent event. Representing more complex patterns involving correlations between multiple edges is very difficult. For example, if two pages point to the same page, it is more likely that they point to each other as well. Such interactions between many overlapping triples of links do not fit well into the generative framework.

Furthermore, directed models such as Bayesian networks and PRMs are usually trained to optimize the joint probability of the labels and other attributes, while the goal of classification is a discriminative model of labels given the other attributes. The advantage of training a model only to discriminate between labels is that it does not have to trade off between classification accuracy and modeling the joint distribution over non-label attributes. In many cases, discriminatively trained models are more robust to violations of independence assumptions and achieve higher classification accuracy than their generative counterparts.

## 3 Undirected Models for Classification

As discussed, our approach to the collective classification task is based on the use of undirected graphical models. We begin by reviewing *Markov networks*, a "flat" undirected model. We then discuss how Markov networks can be extended to the relational setting.

**Markov networks.** We use $\mathbf{V}$ to denote a set of discrete random variables and $\mathbf{v}$ an assignment of values to $\mathbf{V}$. A Markov network for $\mathbf{V}$ defines a joint distribution over $\mathbf{V}$. It consists of a qualitative component, an undirected dependency graph, and a quantitative component, a set of parameters associated with the graph. For a graph $G$, a *clique* is a set of nodes $\mathbf{V}_c$ in $G$, not necessarily maximal, such that each $V_i, V_j \in \mathbf{V}_c$ are connected by an edge in $G$. Note that a single node is also considered a clique.

**Definition 1:** Let $G = (\mathbf{V}, E)$ be an undirected graph with a set of cliques $C(G)$. Each $c \in C(G)$ is associated with a set of nodes $\mathbf{V}_c$ and a *clique potential* $\phi_c(\mathbf{V}_c)$, which is a non-negative function defined on the joint domain of $\mathbf{V}_c$. Let $\Phi = \{\phi_c(\mathbf{V}_c)\}_{c \in C(G)}$. The Markov net $(G, \Phi)$ defines the distribution $P(\mathbf{v}) = \frac{1}{Z} \prod_{c \in C(G)} \phi_c(\mathbf{v}_c)$, where $Z$ is the *partition function* — a normalization constant given by $Z = \sum_{\mathbf{v}'} \prod \phi_c(\mathbf{v}'_c)$. ∎

Each potential $\phi_c$ is simply a table of values for each assignment $\mathbf{v}_c$ that defines a "compatibility" between values of variables in the clique. The potential is often represented by a log-linear combination of a small set of indicator functions, or *features*, of the form $f(\mathbf{V}_c) \equiv \delta(\mathbf{V}_c = \mathbf{v}_c)$. In this case, the potential can be more conveniently rep-

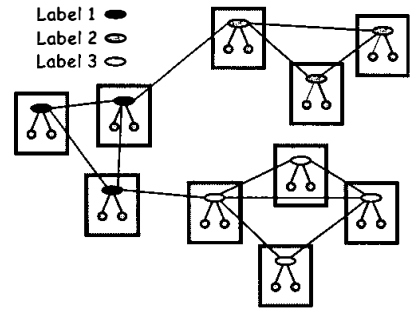

Figure 1: An unrolled Markov net over linked documents. The links follow a common pattern: documents with the same label tend to link to each other more often.

resented in log-linear form:

$$\phi_c(\mathbf{v}_c) = \exp\{\sum_i w_i f_i(\mathbf{v}_c)\} = \exp\{\mathbf{w}_c \cdot \mathbf{f}_c(\mathbf{v}_c)\} .$$

Hence we can write:

$$\log P(\mathbf{v}) = \sum_c \mathbf{w}_c \cdot \mathbf{f}_c(\mathbf{v}_c) - \log Z = \mathbf{w} \cdot \mathbf{f}(\mathbf{v}) - \log Z$$

where $\mathbf{w}$ and $\mathbf{f}$ are the vectors of all weights and features.

For classification, we are interested in constructing discriminative models using *conditional Markov nets* which are simply Markov networks renormalized to model a conditional distribution.

**Definition 2:** Let $\mathbf{X}$ be a set of random variables on which we condition and $\mathbf{Y}$ be a set of target (or label) random variables. A *conditional Markov network* is a Markov network $(G, \Phi)$ which defines the distribution $P(\mathbf{y} \mid \mathbf{x}) = \frac{1}{Z(\mathbf{x})} \prod_{c \in C(G)} \phi_c(\mathbf{x}_c, \mathbf{y}_c)$, where $Z(\mathbf{x})$ is the partition function, now dependent on $\mathbf{x}$: $Z(\mathbf{x}) = \sum_{\mathbf{y}'} \prod \phi_c(\mathbf{x}_c, \mathbf{y}'_c)$. ∎

Logistic regression, a well-studied statistical model for classification, can be viewed as the simplest example of a conditional Markov network. In standard form, for $Y = \pm 1$ and $\mathbf{X} \in \{0,1\}^n$ (or $\mathbf{X} \in \Re^n$), $P(y \mid \mathbf{x}) = \frac{1}{Z(\mathbf{x})} \exp\{y\mathbf{w} \cdot \mathbf{x}\}$. Viewing the model as a Markov network, the cliques are simply the edges $c_k = \{X_k, Y\}$ with potentials $\phi_k(x_k, y) = \exp\{yw_k x_k\}$.

**Relational Markov Networks.** We now extend the framework of Markov networks to the relational setting. A *relational Markov network (RMN)* specifies a conditional distribution over all of the labels of all of the entities in an instantiation given the relational structure and the content attributes. (We provide the definitions directly for the conditional case, as the unconditional case is a special case where the set of content attributes is empty.) Roughly speaking, it specifies the cliques and potentials between attributes of related entities at a template level, so a single model provides a coherent distribution for any collection of instances from the schema.

For example, suppose that pages with the same label tend to link to each other, as in Fig. 1. We can capture this correlation between labels by introducing, for each link, a clique between the labels of the source and the target page. The potential on the clique will have higher values for assignments that give a common label to the linked pages.

To specify what cliques should be constructed in an instantiation, we will define a notion of a *relational clique template*. A relational clique template specifies tuples of variables in the instantiation by using a relational query language. For our link example, we can write the template as a kind of SQL query:

SELECT doc1.Category, doc2.Category
FROM Doc doc1, Doc doc2, Link link
WHERE link.From = doc1.Key and link.To = doc2.Key

Note the three clauses that define a query: the FROM clause specifies the cross product of entities to be filtered by the WHERE clause and the SELECT clause picks out the attributes of interest. Our definition of clique templates contains the corresponding three parts.

**Definition 3:** A *relational clique template* $C = (\mathbf{F}, \mathbf{W}, \mathbf{S})$ consists of three components:

- $\mathbf{F} = \{F_i\}$ — a set of entity variables, where an entity variable $F_i$ is of type $E(F_i)$.
- $\mathbf{W}(\mathbf{F}.\mathbf{R})$ — a boolean formula using conditions of the form $F_i.R_j = F_k.R_l$.
- $\mathbf{F}.\mathbf{S} \subseteq \mathbf{F}.\mathbf{X} \cup \mathbf{F}.\mathbf{Y}$ — a selected subset of content and label attributes in $\mathbf{F}$. ∎

For the clique template corresponding to the SQL query above, $\mathbf{F}$ consists of *doc1*, *doc2* and *link* of types Doc, Doc and Link, respectively. $\mathbf{W}(\mathbf{F}.\mathbf{R})$ is $link.From = doc1.Key \wedge link.To = doc2.Key$ and $\mathbf{F}.\mathbf{S}$ is $doc1.Category$ and $doc2.Category$.

A clique template specifies a set of cliques in an instantiation $\mathcal{I}$:

$$C(\mathcal{I}) \equiv \{c = \mathbf{f}.\mathbf{S} : \mathbf{f} \in \mathcal{I}(\mathbf{F}) \wedge \mathbf{W}(\mathbf{f}.\mathbf{r})\},$$

where $\mathbf{f}$ is a tuple of entities $\{f_i\}$ in which each $f_i$ is of type $E(F_i)$; $\mathcal{I}(\mathbf{F}) = \mathcal{I}(E(F_1)) \times \ldots \times \mathcal{I}(E(F_n))$ denotes the cross-product of entities in the instantiation; the clause $\mathbf{W}(\mathbf{f}.\mathbf{r})$ ensures that the entities are related to each other in specified ways; and finally, $\mathbf{f}.\mathbf{S}$ selects the appropriate attributes of the entities. Note that the clique template does not specify the nature of the interaction between the attributes; that is determined by the clique potentials, which will be associated with the template.

This definition of a clique template is very flexible, as the WHERE clause of a template can be an arbitrary predicate. It allows modeling complex relational patterns on the instantiation graphs. To continue our webpage example, consider another common pattern in hypertext: links in a webpage tend to point to pages of the same category. This pattern can be expressed by the following template:

SELECT doc1.Category, doc2.Category
FROM Doc doc1, Doc doc2, Link link1, Link link2
WHERE link1.From = link2.From and link1.To = doc1.Key and link2.To = doc2.Key and not doc1.Key = doc2.Key

Depending on the expressive power of our template definition language, we may be able to construct very complex templates that select entire subgraph structures of an instantiation. We can easily represent patterns involving three (or more) interconnected documents without worrying about the acyclicity constraint imposed by directed models. Since the clique templates do not explicitly depend on the identities of entities, the same template can select subgraphs whose structure is fairly different. The RMN allows us to associate the same clique potential parameters with all of the subgraphs satisfying the template, thereby allowing generalization over a wide range of different structures.

**Definition 4:** A *Relational Markov network (RMN)* $\mathcal{M} = (\mathbf{C}, \Phi)$ specifies a set of clique templates $\mathbf{C}$ and corresponding potentials $\Phi = \{\phi_C\}_{C \in \mathbf{C}}$ to define a conditional distribution:

$$P(\mathcal{I}.\mathbf{y} \mid \mathcal{I}.\mathbf{x}, \mathcal{I}.\mathbf{r}) = \frac{1}{Z(\mathcal{I}.\mathbf{x}, \mathcal{I}.\mathbf{r})} \prod_{C \in \mathbf{C}} \prod_{c \in C(\mathcal{I})} \phi_C(\mathcal{I}.\mathbf{x}_c, \mathcal{I}.\mathbf{y}_c)$$

where $Z(\mathcal{I}.\mathbf{x}, \mathcal{I}.\mathbf{r})$ is the normalizing partition function: $Z(\mathcal{I}.\mathbf{x}, \mathcal{I}.\mathbf{r}) = \sum_{\mathcal{I}.\mathbf{y}'} \prod_{C \in \mathbf{C}} \prod_{c \in C(\mathcal{I})} \phi_C(\mathcal{I}.\mathbf{x}_c, \mathcal{I}.\mathbf{y}'_c)$ ∎

Using the log-linear representation of potentials, $\phi_C(\mathbf{V}_C) = \exp\{\mathbf{w}_C \cdot \mathbf{f}_C(\mathbf{V}_C)\}$, we can write

$$\log P(\mathcal{I}.\mathbf{y} \mid \mathcal{I}.\mathbf{x}, \mathcal{I}.\mathbf{r})$$
$$= \sum_{C \in \mathbf{C}} \sum_{c \in C(\mathcal{I})} \mathbf{w}_C \cdot \mathbf{f}_C(\mathcal{I}.\mathbf{x}_c, \mathcal{I}.\mathbf{y}_c) - \log Z(\mathcal{I}.\mathbf{x}, \mathcal{I}.\mathbf{r})$$
$$= \sum_{C \in \mathbf{C}} \mathbf{w}_C \cdot \mathbf{f}_C(\mathcal{I}.\mathbf{x}, \mathcal{I}.\mathbf{y}, \mathcal{I}.\mathbf{r}) - \log Z(\mathcal{I}.\mathbf{x}, \mathcal{I}.\mathbf{r})$$
$$= \mathbf{w} \cdot \mathbf{f}(\mathcal{I}.\mathbf{x}, \mathcal{I}.\mathbf{y}, \mathcal{I}.\mathbf{r}) - \log Z(\mathcal{I}.\mathbf{x}, \mathcal{I}.\mathbf{r})$$

where

$$\mathbf{f}_C(\mathcal{I}.\mathbf{x}, \mathcal{I}.\mathbf{y}, \mathcal{I}.\mathbf{r}) = \sum_{c \in C(\mathcal{I})} \mathbf{f}_C(\mathcal{I}.\mathbf{x}_c, \mathcal{I}.\mathbf{y}_c)$$

is the sum over all appearances of the template $C(\mathcal{I})$ in the instantiation, and $\mathbf{f}$ is the vector of all $\mathbf{f}_C$.

Given a particular instantiation $\mathcal{I}$ of the schema, the RMN $\mathcal{M}$ produces an *unrolled* Markov network over the attributes of entities in $\mathcal{I}$. The cliques in the unrolled network are determined by the clique templates $C$. We have one clique for each $c \in C(\mathcal{I})$, and all of these cliques are associated with the same clique potential $\phi_C$. In our webpage example, an RMN with the link feature described above would define a Markov net in which, for every link between two pages, there is an edge between the labels of these pages. Fig. 1 illustrates a simple instance of this unrolled Markov network.

## 4 Learning the Models

In this paper, we focus on the case where the clique templates are given; our task is to estimate the clique potentials, or feature weights. Thus, assume that we are given a set of clique templates **C** which partially specify our (relational) Markov network, and our task is to compute the weights **w** for the potentials $\Phi$. In the learning task, we are given some training set $D$ where both the content attributes and the labels are observed. Any particular setting for **w** fully specifies a probability distribution $P_\mathbf{w}$ over $D$, so we can use the *likelihood* as our objective function, and attempt to find the weight setting that maximizes the likelihood (ML) of the labels given other attributes. However, to help avoid overfitting, we assume a "shrinkage" prior over the weights (a zero-mean Gaussian), and use maximum a posteriori (MAP) estimation. More precisely, we assume that different parameters are a priori independent and define $p(w_i) = \frac{1}{\sqrt{2\pi\sigma^2}} \exp\left\{-w_i^2/2\sigma^2\right\}$.

Both the ML and MAP objective functions are concave and there are many methods available for maximizing them. Our experience is that conjugate gradient and even simple gradient perform very well for logistic regression (Minka 2000) and relational Markov nets.

**Learning Markov Networks.** We first consider discriminative MAP training in the flat setting. In this case $D$ is simply a set of IID instances; let $d$ index over all labeled training data $D$. The discriminative likelihood of the data is $\prod_d P_\mathbf{w}(y_d \mid \mathbf{x}_d)$. We introduce the parameter prior, and maximize the log of the resulting MAP objective function:

$$L(\mathbf{w}, D) = \sum_{d \in D} \left(\mathbf{w} \cdot \mathbf{f}(\mathbf{x}_d, y_d) - \log Z(\mathbf{x}_d)\right) - \frac{\mathbf{w} \cdot \mathbf{w}}{2\sigma^2} + C \ .$$

The gradient of the objective function is computed as:

$$\nabla L(\mathbf{w}, D) = \sum_{d \in D} \left(\mathbf{f}(\mathbf{x}_d, y_d) - E_{P_\mathbf{w}}[\mathbf{f}(\mathbf{x}_d, Y_d)]\right) - \frac{\mathbf{w}}{\sigma^2} \ .$$

The last term is the shrinking effect of the prior and the other two terms are the difference between the expected feature counts and the empirical feature counts, where the expectation is taken relative to $P_\mathbf{w}$:

$$E_{P_\mathbf{w}}[\mathbf{f}(\mathbf{x}_d, Y_d)] = \sum_{y'} \mathbf{f}(\mathbf{x}_d, y'_d) P_\mathbf{w}(y'_d \mid \mathbf{x}_d) \ .$$

Thus, ignoring the effect of the prior, the gradient is zero when empirical and expected feature counts are equal.[1] The prior term gives the smoothing we expect from the prior: small weights are preferred in order to reduce overfitting. Note that the sum over $y'$ is just over the possible categorizations for one data sample every time.

**Learning RMNs.** The analysis for the relational setting is very similar. Now, our data set $D$ is actually a single instantiation $\mathcal{I}$, where the same parameters are used multiple times — once for each different entity that uses a feature. A particular choice of parameters **w** specifies a particular RMN, which induces a probability distribution $P_\mathbf{w}$ over the unrolled Markov network. The product of the likelihood of $\mathcal{I}$ and the parameter prior define our objective function, whose gradient $\nabla L(\mathbf{w}, \mathcal{I})$ again consists of the empirical feature counts minus the expected features counts and a smoothing term due to the prior:

$$\mathbf{f}(\mathcal{I}.\mathbf{y}, \mathcal{I}.\mathbf{x}, \mathcal{I}.\mathbf{r}) - E_\mathbf{w}[\mathbf{f}(\mathcal{I}.\mathbf{Y}, \mathcal{I}.\mathbf{x}, \mathcal{I}.\mathbf{r})] - \frac{\mathbf{w}}{\sigma^2}$$

where the expectation $E_{P_\mathbf{w}}[\mathbf{f}(\mathcal{I}.\mathbf{Y}, \mathcal{I}.\mathbf{x}, \mathcal{I}.\mathbf{r})]$ is

$$\sum_{\mathcal{I}.\mathbf{y}'} \mathbf{f}(\mathcal{I}.\mathbf{y}', \mathcal{I}.\mathbf{x}, \mathcal{I}.\mathbf{r}) P_\mathbf{w}(\mathcal{I}.\mathbf{y}' \mid \mathcal{I}.\mathbf{x}, \mathcal{I}.\mathbf{r}) \ .$$

This last formula reveals a key difference between the relational and the flat case: the sum over $\mathcal{I}.\mathbf{y}'$ involves the exponential number of assignments to all the label attributes in the instantiation. In the flat case, the probability decomposes as a product of probabilities for individual data instances, so we can compute the expected feature count for each instance separately. In the relational case, these labels are correlated — indeed, this correlation was our main goal in defining this model. Hence, we need to compute the expectation over the joint assignments to all the entities together. Computing these expectations over an exponentially large set is the expensive step in calculating the gradient. It requires that we run inference on the unrolled Markov network.

**Inference in Markov Networks.** The inference task in our conditional Markov networks is to compute the posterior distribution over the label variables in the instantiation given the content variables. Exact algorithms for inference in graphical models can execute this process efficiently for specific graph topologies such as sequences, trees and other low treewidth graphs. However, the networks resulting from domains such as our hypertext classification task are very large (in our experiments, they contain tens of thousands of nodes) and densely connected. Exact inference is completely intractable in these cases.

We therefore resort to approximate inference. There is a wide variety of approximation schemes for Markov networks. We chose to use *belief propagation* for its simplicity and relative efficiency and accuracy. Belief Propagation (BP) is a local message passing algorithm introduced by Pearl (1988). It is guaranteed to converge to the correct marginal probabilities for each node only for singly connected Markov networks. However, recent analysis (Yedidia *et al.* 2000) provides some theoretical justification. Empirical results (Murphy *et al.* 1999) show that it often converges in general networks, and when it does, the marginals are a good approximation to the correct posteriors. As our results in Section 5 show, this approach works well in our domain. We refer the reader to Yedidia *et al.* for a detailed description of the BP algorithm.

## 5 Experiments

We tried out our framework on the *WebKB* dataset (Craven *et al.* 1998), which is an instance of our hypertext exam-ple. The data set contains webpages from four different Computer Science departments: Cornell, Texas, Washington and Wisconsin. Each page has a label attribute, representing the type of webpage which is one of *course, faculty, student, project* or *other*. The data set is problematic in that the category *other* is a grab-bag of pages of many different types. The number of pages classified as *other* is quite large, so that a baseline algorithm that simply always selected *other* as the label would get an average accuracy of 75%. We could restrict attention to just the pages with the four other labels, but in a relational classification setting, the deleted webpages might be useful in terms of their interactions with other webpages. Hence, we compromised by eliminating all *other* pages with fewer than three outlinks, making the number of *other* pages commensurate with the other categories.[2] For each page, we have access to the entire html of the page and the links to other pages. Our goal is to collectively classify webpages into one of these five categories. In all of our experiments, we learn a model from three schools and test the performance of the learned model on the remaining school, thus evaluating the generalization performance of the different models.

---

[1] The solution of maximum likelihood estimation with log-linear models is actually also the solution to the dual problem of maximum entropy estimation with constraints that empirical and expected feature counts must be equal (Della Pietra *et al.* 1997).

Unfortunately, we cannot directly compare our accuracy results with previous work because different papers use different subsets of the data and different training/test splits. However, we compare to standard text classifiers such as Naive Bayes, Logistic Regression, and Support Vector Machines, which have been demonstrated to be successful on this data set (Joachims 1999).

**Flat Models.** The simplest approach we tried predicts the categories based on just the text content on the webpage. The text of the webpage is represented using a set of binary attributes that indicate the presence of different words on the page. We found that stemming and feature selection did not provide much benefit and simply pruned words that appeared in fewer than three documents in each of the three schools in the training data. We also experimented with incorporating meta-data: words appearing in the title of the page, in anchors of links to the page and in the last header before a link to the page (Yang *et al.* 2002). Note that meta-data, although mostly originating from pages linking into the considered page, are easily incorporated as features, i.e. the resulting classification task is still flat feature-based classification. Our first experimental setup compares three well-known text classifiers — Naive Bayes, linear support vector machines [3] (Svm), and logistic regression (Logistic) — using words and meta-words. The results, shown in Fig. 2(a), show that the two discriminative approaches outperform Naive Bayes. Logistic and Svm give very similar results. The average error over the 4 schools was reduced by around 4% by introducing the meta-data attributes.

---

[2]The resulting category distribution is: course (237), faculty (148), other (332), research-project (82) and student (542). The number of remaining pages for each school are: Cornell (280), Texas (292), Washington (315) and Wisconsin (454). The number of links for each school are: Cornell (574), Texas (574), Washington (728) and Wisconsin (1614).

[3]We trained one-against-others Svm for each category and during testing, picked the category with the largest margin.

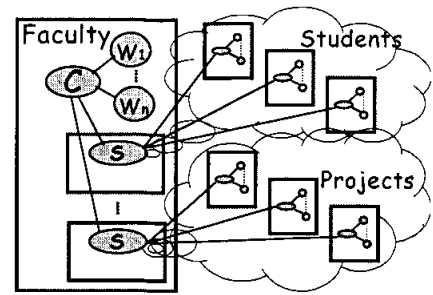

Figure 3: An illustration of the Section model.

**Relational Models.** Incorporating meta-data gives a significant improvement, but we can take additional advantage of the correlation in labels of related pages by classifying them collectively. We want to capture these correlations in our model and use them for transmitting information between linked pages to provide more accurate classification. We experimented with several relational models. Recall that logistic regression is simply a flat conditional Markov network. All of our relational Markov networks use a logistic regression model locally for each page.

Our first model captures direct correlations between labels of linked pages. These correlations are very common in our data: courses and research projects almost never link to each other; faculty rarely link to each other; students have links to all categories but mostly courses. The Link model, shown in Fig. 1, captures this correlation through links: in addition to the local bag of words and meta-data attributes, we introduce a relational clique template over the labels of two pages that are linked.

A second relational model uses the insight that a webpage often has internal structure that allows it to be broken up into *sections*. For example, a faculty webpage might have one section that discusses research, with a list of links to all of the projects of the faculty member, a second section might contain links to the courses taught by the faculty member, and a third to his advisees. This pattern is illustrated in Fig. 3. We can view a section of a webpage as a fine-grained version of Kleinberg's hub (Kleinberg 1999) (a page that contains a lot of links to pages of particular category). Intuitively, if we have links to two pages in the same section, they are likely to be on similar topics. To take advantage of this trend, we need to enrich our schema with a new relation Section, with attributes *Key, Doc* (document in which it appears), and *Category*. We also need to add the attribute *Section* to Link to refer to the section it appears in. In the RMN, we have two new relational clique templates. The first contains the label of a section and the label of the page it is on:

SELECT doc.Category, sec.Category
FROM Doc doc, Section sec
WHERE sec.Doc = doc.Key

The second clique template involves the label of the section

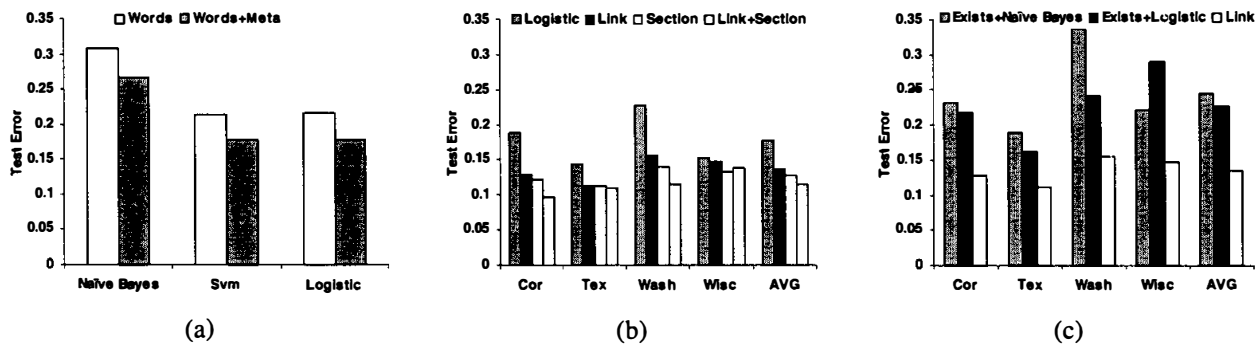

Figure 2: (a) Comparison of Naive Bayes, Svm, and Logistic on WebKB, with and without meta-data features. (Only averages over the 4 schools are shown here.) (b) Flat versus collective classification on WebKB: flat logistic regression with meta-data, and three different relational models: Link, Section, and a combined Section+Link. (c) Comparison of generative and discriminative relational models. Exists+NaiveBayes is completely generative. Exists+Logistic is generative in the links, but locally discriminative in the page labels given the local features (words, meta-words). The Link model is completely discriminative.

containing the link and the label of the target page.

SELECT sec.Category, doc.Category
FROM Section sec, Link link, Doc doc
WHERE link.Sec = sec.Key and link.To = doc.Key

The original dataset did not contain section labels, so we introduced them using the following simple procedure. We defined a section as a sequence of three or more links that have the same path to the root in the html parse tree. In the training set, a section is labeled with the most frequent category of its links. There is a sixth category *none*, assigned when the two most frequent categories of the links are less than a factor of 2 apart. In the entire data set, the breakdown of labels for the sections we found is: *course* (40), *faculty* (24), *other* (187), *research.project* (11), *student* (71) and *none* (17). Note that these labels are hidden in the test data, so the learning algorithm now also has to learn to predict section labels. Although not our final aim, correct prediction of section labels is very helpful. Words appearing in the last header before the section are used to better predict the section label by introducing a clique over these words and section labels.

We compared the performance of Link, Section and Section+Link (a combined model which uses both types of cliques) on the task of predicting webpage labels, relative to the baseline of flat logistic regression with meta-data. Our experiments used MAP estimation with a Gaussian prior on the feature weights with standard deviation of 0.3. Fig. 2(b) compares the average error achieved by the different models on the four schools, training on three and testing on the fourth. We see that incorporating any type of relational information consistently gives significant improvement over the baseline model. The Link model incorporates more relational interactions, but each is a weaker indicator. The Section model ignores links outside of coherent sections, but each of the links it includes is a very strong indicator. In general, we see that the Section models performs slightly better. The joint model is able to combine benefits from both and generally outperforms all of the other models. The only exception is for the task of classifying the Wisconsin data. In this case, the joint Section+Link model contains many links, as well as some large tightly connected loops, so belief propagation did not converge for a subset of nodes. Hence, the results of the inference, which was stopped at a fixed arbitrary number of iterations, were highly variable and resulted in lower accuracy.

**Discriminative vs Generative.** Our last experiment illustrates the benefits of discriminative training in relational classification. We compared three models. The Exists+Naive Bayes model is a completely generative model proposed by Getoor *et al.* (2001). At each page, a naive Bayes model generates the words on a page given the page label. A separate generative model specifies a probability over the existence of links between pages conditioned on both pages' labels. We can also consider an alternative Exists+Logistic model that uses a discriminative model for the connection between page label and words — i.e. uses logistic regression for the conditional probability distribution of page label given words. This model has equivalent expressive power to the naive Bayes model but is discriminatively rather than generatively trained. Finally, the Link model is a fully discriminative (undirected) variant we have presented earlier, which uses a discriminative model for the label given both words and link existence. The results, shown in Fig. 2(c), show that discriminative training provides a significant improvement in accuracy: the Link model outperforms Exists+Logistic which in turn outperforms Exists+Naive Bayes.

As illustrated in Table 1, the gain in accuracy comes at some cost in training time: for the generative models, parameter estimation is closed form while the discriminative models are trained using conjugate gradient, where each iteration requires inference over the unrolled RMN. On the other hand, both types of models require inference when the model is used on new data; the generative model con-structs a much larger, fully-connected network, resulting in significantly longer testing times. We also note that the situation changes if some of the data is unobserved in the training set. In this case, generative training also require an iterative procedure (such as EM) where each iteration uses the significantly more expressive inference.

|  | Links | Links+Section | Exists+NB |
|---|---|---|---|
| Training | 1530 | 6060 | 1 |
| Testing | 7 | 10 | 100 |

Table 1: Average train/test running times (seconds). All runs were done on a 700Mhz Pentium III. Training times are averaged over four runs on three schools each. Testing times are averaged over four runs on one school each.

## 6 Discussion and Conclusions

In this paper, we propose a new approach for classification in relational domains. Our approach provides a coherent probabilistic foundation for the process of collective classification, where we want to classify multiple entities, exploiting the interactions between their labels. We have shown that we can exploit a very rich set of relational patterns in classification, significantly improving the classification accuracy over standard flat classification.

In some cases, we can incorporate relational features into standard flat classification. For example, when classifying papers into topics, it is possible to simply view the presence of particular citations as atomic features. However, this approach is limited in cases where some or even all of the relational features that occur in the test data are not observed in the training data. In our WebKB example, there is no overlap between the webpages in the different schools, so we cannot learn anything from the training data about the significance of a hyperlink to/from a particular webpage in the test data. Incorporating basic features (e.g., words) from the related entities can aid in classification, but cannot exploit the strong correlation between the *labels* of related entities that RMNs capture.

Our results in this paper are only a first step towards understanding the power of relational classification. On the technical side, we can gain significant power from introducing hidden variables (that are not observed even in the training data), such as the degree to which a webpage is an authority (Kleinberg 1999). Furthermore, as we discussed, there are many other types of relational patterns that we can exploit. We can also naturally extend the proposed models to predict relations between entities, for example, advisor-advisee, instructor-course or project-member.

Hypertext is the most easily available source of structured data, however, RMNs are generally applicable to any relational domain. In particular, social networks provide extensive information about interactions among people and organizations. RMNs offer a principled method for learning to predict communities of and hierarchical structure between people and organizations based on both the local attributes and the patterns of static and dynamic interaction. Given the wealth of possible patterns, it is particularly interesting to explore the problem of inducing them automatically. We intend to explore this topic in future work.

**Acknowledgments.** This work was supported by ONR Contract F3060-01-2-0564-P00002 under DARPA's EELD program. P. Abbeel was also supported by a Siebel Graduate Fellowship.

## References

S. Chakrabarti, B. Dom, and P. Indyk. Enhanced hypertext categorization using hyperlinks. In *Proc. of ACM SIGMOD98*, pages 307–318, 1998.

M. Craven, D. DiPasquo, D. Freitag, A. McCallum, T. Mitchell, K. Nigam, and S. Slattery. Learning to extract symbolic knowledge from the world wide web. In *Proc AAAI98*, pages 509–516, 1998.

S. Della Pietra, V. Della Pietra, and J. Lafferty. Inducing features of random fields. *IEEE Trans. on Pattern Analysis and Machine Intelligence*, 19(4):380-393, 1997.

L. Egghe and R. Rousseau. *Introduction to Informetrics*. Elsevier, 1990.

N. Friedman, L. Getoor, D. Koller, and A. Pfeffer. Learning probabilistic relational models. In *Proc. IJCAI99*, pages 1300–1309, Stockholm, Sweden, 1999.

L. Getoor, E. Segal, B. Taskar, and D. Koller. Probabilistic models of text and link structure for hypertext classification. In *Proc. IJCAI01 Workshop on Text Learning: Beyond Supervision*, Seattle, Wash., 2001.

T. Joachims. Transductive inference for text classification using support vector machines. In *Proc. ICML99*, pages 200-209. Morgan Kaufmann Publishers, San Francisco, US, 1999.

J. M. Kleinberg. Authoritative sources in a hyperlinked environment. *Journal of the ACM*, 46(5):604-632, 1999.

D. Koller and A. Pfeffer. Probabilistic frame-based systems. In *Proc. AAAI98*, pages 580-587, Madison, Wisc., 1998.

J. Lafferty, A. McCallum, and F. Pereira. Conditional random fields: Probabilistic models for segmenting and labeling sequence data. In *Proc. ICML01*, 2001.

T. Minka. Algorithms for maximum-likelihood logistic regression. http://lib.stat.cmu.edu/ minka/papers/logreg.html, 2000.

K. P. Murphy, Y. Weiss, and M. I. Jordan. Loopy belief propagation for approximate inference: an empirical study. In *Proc. UAI99*, pages 467–475, 1999.

J. Neville and D. Jensen. Iterative classification in relational data. In *Proc. AAAI-2000 Workshop on Learning Statistical Models from Relational Data*, pages 13–20, 2000.

J. Pearl. *Probabilistic Reasoning in Intelligent Systems*. Morgan Kaufmann, San Francisco, 1988.

S. Slattery and T. Mitchell. Discovering test set regularities in relational domains. In *Proc. ICML00*, pages 895–902, 2000.

B. Taskar, E. Segal, and D. Koller. Probabilistic classification and clustering in relational data. In *Proc. IJCAI01*, pages 870–876, Seattle, Wash., 2001.

V.N. Vapnik. *The Nature of Statistical Learning Theory*. Springer-Verlag, New York, New York, 1995.

Y. Yang, S. Slattery, and R. Ghani. A study of approaches to hypertext categorization. *Journal of Intelligent Information Systems*, 18(2), 2002.

J. Yedidia, W. Freeman, and Y. Weiss. Generalized belief propagation. In *NIPS*, pages 689–695, 2000.